\documentclass[letterpaper, 10 pt, conference]{ieeeconf}  % 

\IEEEoverridecommandlockouts                              % 
\usepackage[utf8]{inputenc}
\usepackage{graphicx}
\usepackage{amsmath} % assumes amsmath package installed
\usepackage{amssymb}  % assumes amsmath package installed

\usepackage{booktabs} 
\usepackage{hyperref}

\usepackage{fdsymbol}
\usepackage{float}
\usepackage[all]{xy}
\usepackage{algorithmic}
\usepackage{multirow}
\usepackage{algorithm}
\usepackage{lipsum}
\usepackage{xcolor}
\usepackage[space]{cite}
\usepackage{threeparttable}

\newcommand{\revise}[1]{{\color{black}{#1}}}

\title{\LARGE \bf
Dynamic Causal Graph Convolutional Network for Traffic Prediction
}
\author{Junpeng Lin$^{1}$, Ziyue Li$^{2}$, Zhishuai Li$^{3}$, Lei Bai$^{4}$, Rui Zhao$^{5}$ and Chen Zhang$^{1}$% <-this % stops a space
\thanks{$^{1}$Junpeng Lin and Chen Zhang are with the Department of Industrial Engineering, Tsinghua University, Beijing, China (Email: linjp22@mails.tsinghua.edu.cn; zhangchen01@tsinghua.edu.cn)}
\thanks{$^{2}$Ziyue Li is with University of Cologne, 50923 Cologne, NRW, Germany (Email: zlibn@wiso.uni-koeln.de)}
\thanks{$^{3}$Zhishuai Li is with SenseTime Research, Shanghai, China (Email: lizhishuai@sensetime.com)}
\thanks{$^{4}$Lei Bai is with The Shanghai AI Laboratory, Shanghai, China (Email: baisanshi@gmail.com)}
\thanks{$^{5}$Rui Zhao is with SenseTime Research and Qing Yuan Research Institute of Shanghai Jiao Tong University, Shanghai, China (Email: zhaorui@sensetime.com)}
}

\begin{document}

\maketitle
\thispagestyle{empty}
\pagestyle{empty}

% NOTES ON BUG
% 1. notations:
%       - input and output sequences length and dimension of X

%%%%%%%%%%%%%%%%%%%%%%%%%%%%%%%%%%%%%%%%%%%%%%%%%%%%%%%%%%%%%%%%%%%`%%%%%%%%%%%%%
\begin{abstract}

% Modeling complex spatiotemporal dependencies for correlated traffic series is critical for traffic forecasting. Recent works improve prediction performance by developing complicated neural network structures to extract spatiotemporal correlations, whose effectiveness depends on the preciseness of well-defined graph structures representing the spatial topology of the traffic network. We propose to embed time-varying dynamic Bayesian graphs in traffic forecasting to represent the fine spatiotemporal topology of traffic data, to which graph convolution networks are applied to produce traffic forecasts. A deep learning based module is designed as a hyper-network to generate stepwise dynamic causal graphs capable of modeling nonlinear traffic propagation patterns, which is more efficient compared to classical causal learning methods. Experiments on a real traffic dataset demonstrate the superior predictive performance of the proposed method.

Modeling complex spatiotemporal dependencies in correlated traffic series is essential for traffic prediction. While recent works have shown improved prediction performance by using neural networks to extract spatiotemporal correlations, their effectiveness depends on the quality of the graph structures used to represent the spatial topology of the traffic network. In this work, we propose a novel approach for traffic prediction that embeds time-varying dynamic Bayesian network to capture the fine spatiotemporal topology of traffic data. We then use graph convolutional networks to generate traffic forecasts. To enable our method to efficiently model nonlinear traffic propagation patterns, we develop a deep learning-based module as a hyper-network to generate stepwise dynamic causal graphs. Our experimental results on a real traffic dataset demonstrate the superior prediction performance of the proposed method. The code is available at \url{https://github.com/MonBG/DCGCN}.

\end{abstract}

%%%%%%%%%%%%%%%%%%%%%%%%%%%%%%%%%%%%%%%%%%%%%%%%%%%%%%%%%%%%%%%%%%%%%%%%%%%%%%%%
\section{INTRODUCTION}
% 1 page

% problem background

% from classical statistical methods to deep-based methods

% key challenges of traffic prediction, instead of construct a complex graph convolution structure, we make efforts to construct a finer spatial topology

% contributions
% 1. deep-based tvdbn generation
% 2. application of tvdbn in traffic prediction
% 3. suitable sequential training strategy
% 4. experiments with sota methods, better performance

Urban transportation plays a crucial role in modern life. With the rapid development of urbanization, the fast growing urban population and the number of cars bring great pressure to urban transportation system. In this situation, traffic prediction can help understand traffic propagation patterns and serves as the cornerstone of traffic management. 
%Traffic forecasting is a challenging task due to the complicated spatial and temporal dependencies in traffic data, i.e., correlations between multiple traffic series from different locations and between different timesteps within a single traffic series, respectively. 

In the last decades, thousands of data-driven methods for traffic prediction have been proposed. Classical statistical methods, e.g., vector auto-regression (VAR) \cite{zivot2006vector}, auto-regressive integrated moving average (ARIMA) \cite{box2015time}, are pioneering works used to model traffic series. Later,
%However, these methods fail when handling complex traffic data that exhibit nonlinear spatiotemporal dependencies.
deep learning based methods \cite{tedjopurnomo2020survey} began to be increasingly popular, and achieved the overwhelming results. Most state-of-the-art deep learning methods integrate graph convolution network (GCN) \cite{kipf2016semi} into recurrent neural networks (RNNs) \cite{li2017diffusion,bai2020adaptive} or 1-D convolution neural networks (CNNs) on the time axis \cite{yu2017spatio,guo2019attention} to model spatiotemporal dependencies of traffic data. However, the effectiveness of GCN generally depends on the quality of the graph adjacency matrix, which is expected to fully capture the spatial characteristics of the real traffic network. Some popularly used graph adjacency matrices include traffic road topology \cite{huang2020lsgcn,ziyue2021tensor,li2022individualized,li2020tensor,mao2022jointly}, traffic data correlation \cite{yao2018deep, tsung2020discussion,li2022profile,li2020long,wang2023correlated}, etc. 

In recent years, casual discovery emerges and is being explored in many machine learning applications. Casual discovery aims to analyze causal relationships behind statistical correlations of different variables and facilitate better machine learning. \revise{Typical approaches to incorporate causal discovery include encoding features from domain-specific causal models as input to downstream tasks \cite{wang2022causalgnn} and learning the structure of causal relationships between features for graph-based models \cite{luan2022traffic, lan2023mm}.} As a powerful graph-based tool for modeling directed causal relationships between variables, Bayesian network (BN) %\revise{\sout{, which is also commonly called directed acyclic graphs (DAGs),}}
is being applied in traffic prediction \cite{luan2022traffic}. By representing the traffic propagation trend with directed links, BN shows its potential for explicitly modeling traffic propagation pattern \cite{liu2011discovering}. Furthermore, BN could be extended to dynamic Bayesian networks (DBNs) \cite{murphy2002dynamic} to capture temporal dependencies in traffic series. 

A limitation of current casual-embedded traffic prediction models is the assumption of stationary temporal dependencies. However, in reality, the dependencies of traffic data in different places do change over time. Recent works \cite{li2021dynamic} based on GCN have proposed to generate a \revise{time-invariant} adaptive graph from trainable node embeddings during model training. However, to our best knowledge, there is no work to incorporate causal learning into adaptive spatial topology construction, which is expected to learn time-varying graphs with good explainability for traffic prediction. 
%They require extensive expertise to create and do not adapt to changing spatial topology, resulting in poor performance in modeling spatiotemporal dependencies. To overcome this limitation, recent works propose to generate a time-invariant adaptive graph from trainable node embeddings during model training, forcing the graph structure to adapt to the structure of the model and input data. However, the adaptive graph learned from model training lacks interpretability and is difficult to gain insight into traffic propagation patterns.
 %Moreover, BN could be extended to dynamic Bayesian networks (DBNs) to capture temporal dependencies and adapt to non-stationary traffic series by learning a time-varying DBN (TVDBN).
 
In this work, we propose a novel causal-embedded approach for traffic prediction. It represents the spatiotemporal traffic network topology using a time-varying DBN (TVDBN), which is designed to adapt to the time-varying traffic propagation patterns by learning DBNs step by step. The learned TVDBN is able to summarize the dynamic spatiotemporal dependencies between nodes. Built upon it, graph convolution is applied to capture spatial dependencies for traffic prediction. We propose a complete deep learning based causal structure learning module that serves as a pre-trained hyper-network to generate the graphs of the TVDBN. The predefined distance graph is further incorporated into graph generator and traffic prediction module as additional prior information to improve the performance. The contributions of our work are summarized as follows:
\begin{itemize}
        % \item We conduct a TVDBN based on GCN and RNN to describe the time-varying causal relationships between traffic data in different places. The augmented Lagrange method is applied for graph generator training to ensure the acyclicity of TVDBN. 
        \item We propose an approach based on GCN and RNN to learn a TVDBN that describes the time-varying causal relationships between different locations in traffic network. The augmented Lagrange method is applied for model training to ensure the acyclicity of the TVDBN. 
        \item To our knowledge, we are the first to learn the structure of TVDBN using deep learning. Compared with traditional causal structure learning methods by machine learning or statistics, our method can capture nonlinear causal relationships between nodes more flexibly. 
        %Unlike previous works, we use neural networks to model the generation mechanism of adjacency matrices instead of directly setting them as learnable parameters. This drastically reduces the time required to learn the causal structure when handling unknown traffic data.
        \item We uses the learned TVDBN as dynamic causal adjacency graphs in the downstream GCN-based traffic prediction module. Curriculum learning is used in training to achieve better performance. The training of TVDBN structure learning and traffic prediction modules are conducted separately and sequentially, which improves performance and reduces the training time. 
        \item We conduct experiments on a widely used traffic benchmark dataset METR-LA to evaluate the traffic prediction performance of the proposed model. The results show that our model can outperform the classical and state-of-the-art baselines in both short-term and long-term prediction performance.
\end{itemize}

\section{RELATED WORK}
% 1 page

\subsection{Deep Learning Based Traffic Prediction}
% outline 
% 1. traditional method, MA, VAR, ARIMA
% 2. Deep-based method, including baselines
The main challenge of traffic prediction is capturing spatiotemporal dependencies in traffic data. This involves three tasks including spatial topology construction, spatial dependency modeling and temporal dependency modeling \cite{li2021dynamic}.
% , which helps to distinguish between different Deep-based traffic prediction methods.

\textbf{Spatial topology construction} aims to summarize the structural information of the traffic network into well-defined graph data structures, which are then used to extract spatial dependencies.
% describe the traffic network that contains the structural information needed to extract spatial dependencies. 
CNN-based methods \cite{yao2018deep,lin2019deepstn+} were first proposed to divide maps into equally sized grids as images, and used convolution to describe correlations between neighbouring girds. Later to better process non-Euclidean correlations, graph neural networks (GNNs) provide a more flexible representation of the traffic network. For GNNs, it is crucial to construct a suitable adjacency matrix. The most common way is to predefine a proximity metric between pairs of nodes, such as geographic distance or connectivity \cite{li2017diffusion,yu2017spatio,zheng2020gman}. However, the predefined adjacency matrix is static and has limited ability to describe highly dynamic spatial correlations between traffic data. To address this, some methods set the adjacency matrix as learnable parameters \cite{zhang2020spatio} or generate it from learnable node embeddings \cite{bai2020adaptive, wu2020connecting}, consequently allowing the graph to adapt to the actual data. The Dynamic Graph Convolutional Recurrent Network (DGCRN) \cite{li2021dynamic} goes further by generating a self-adaptive dynamic adjacency matrix at each time step. While this approach offers high representational flexibility, the resulting dynamic graphs lack interpretability in practical applications, limiting their generalization to other traffic analysis scenarios.

\textbf{Spatial dependency modeling} targets at mining correlations between spatial nodes from the constructed spatial topology. For grid-based spatial modeling, CNN-based methods \cite{yao2018deep,lin2019deepstn+} apply 2-D convolution to capture spatial dependencies in Euclidean data. For graphs represented spatial modeling, most state-of-the-art methods use GNNs to process spatial information by aggregating messages from neighboring nodes through graph convolution or attention mechanisms \cite{li2017diffusion,yu2017spatio,zheng2020gman}. Recent works propose more complicated graph convolution modules to enhance the expressive power and address the over-smoothing issue of GCN \cite{wu2020connecting}. 
% One such example is mix-hop propagation used in MTGNN \cite{wu2020connecting}, which introduces an information selection step and skip connections to mitigate the over-smoothing issue.

\textbf{Temporal dependency modeling} is critical in time series analysis. Recurrent neural networks are widely used to capture temporal dependencies between parts of the sequence. 
%The design of recurrent unit provides flexibility for model structure at the expense of time due to looping computations on the time axis. 
GRU \cite{cho2014learning} and LSTM \cite{hochreiter1997long} are further proposed to improve the ability to model long-term dependencies by introducing a gate mechanism to control the ratio of retaining long-term information. In traffic prediction, to incorporate spatial information, an intuitive approach is to use the outputs of spatial modules as input to RNN \cite{zhao2019t}. Some works also modify the computation of gates to include graph convolution in GRU and LSTM \cite{li2017diffusion, bai2020adaptive}. To reduce the computational cost of RNN, CNN can be used in modeling temporal dependencies with 1-D convolution on the time axis \cite{yu2017spatio}. Recently, attention mechanisms are also applied for modeling long-term dependencies, e.g., ASTGCN \cite{guo2019attention} and GMAN \cite{zheng2020gman}. 
% For example, ASTGCN \cite{guo2019attention, zheng2020gman} combines temporal attention with 1-D convolution to achieve finer temporal feature extraction.

\subsection{Bayesian Network}
% outline
% - meaning of BN
% - DAG structure learning: heuristics or statistical optimization
% - TVDBN structure learning

As an efficient method for modeling directed causal relationships between variables, Bayesian networks (BNs) %\revise{\sout{, aka. directed acyclic graphs (DAGs),}} 
have been applied in many applications \cite{hossain2019framework, tsagris2021new}. \revise{BN is a probabilistic graphical model that takes the form of a directed acyclic graphs (DAG).} In a BN, nodes represent variables, and directed edges represent casual dependencies between nodes. By learnisng the edges and parameters of the BN, the joint distribution of all the variables can be analyzed. To further capture causal relationships in time series, dynamic Bayesian networks (DBNs) \cite{murphy2002dynamic} extend this to multivariate time series data by accounting for causality between variables across and within time steps. By further assuming the causality structure changes over time, time-varying DBN (TVDBN) is further proposed \cite{robinson2008non}. In particular, different types of time-varying patterns include step-wise change, smooth change, etc. As to casual structure learning of BN and its variants, there are two types of methods in general: constraint-based and score-based methods \cite{vowels2022d}. Constraint-based methods \cite{runge2019detecting} rely on the validity of statistical assumptions and score-based methods \cite{zheng2018dags} require an appropriate structure score with an efficient combinatorial structure search strategy. By replacing the acyclicity constraint of BN with a continuous penalty, recent works \cite{zheng2018dags} transform score-based methods into the well-studied continuous optimization problem.

Traffic analysis can benefit from the use of Bayesian networks to reveal traffic propagation patterns, which is an alternative to representing spatial topology. Liu et al.\cite{liu2011discovering} proposed to learn a DBN with tree structure to uncover causal interactions between traffic events. DBGCN \cite{luan2022traffic} applys graph convolution to extract spatiotemporal dependencies from a DBN learned by statistics. However, most existing BN and DBN-based models in traffic analysis learn graphs through statistical methods, which are simple linear models with the assumption of a stationary traffic process. Furthermore, to our best knowledge, there is no work embedding TVDBN into traffic prediction to capture time-varying causal relationships of traffic data. 

\section{PRELIMINARIES}
% 3/4 page

\subsection{Problem Definition}

The task of traffic prediction aims to predict the future traffic variables, such as speed and flow, using historical data. The traffic network is denoted as $\mathcal{G}=(\mathcal{V}, \mathcal{E}, \mathbf{A})$, where $\mathcal{V}$ is a set of $N=|\mathcal{V}|$ nodes representing different locations (e.g., sensors or road segments) in the traffic network, and $\mathcal{E}$ is a set of edges representing the geographic connection between nodes. $\mathbf{A}\in \mathbb{R}^{N\times N}$ is the weighted adjacency matrix corresponding to $\mathcal{E}$ where each element represents the connectivity or proximity between nodes. $\mathbf{A}$ is treated as the prior information about the network structure.

The multivariate traffic series are denoted as a feature tensor $\mathbf{X}\in\mathbb{R}^{T_{\text{in}}\times N\times D}$ of $\mathcal{G}$, where $T_{\text{in}}$ is the length of the historical traffic series and $D$ is the number of features of each node. Let $\mathbf{X}_t\in\mathbb{R}^{N\times D}$ represents the traffic state observed at time $t$ and $\mathbf{X}_{t+1:t+T}\in\mathbb{R}^{T\times N\times D}$ represents the features from time $t+1$ to $t+T$. The traffic prediction problem is to learn a function $f$ to predict the next $T_{\text{out}}$ step traffic conditions $\hat{\mathbf{X}}_{t+1:t+T_\text{out}} \in \mathbb{R}^{T_\text{out}\times N\times F}$  based on the past $T_{\text{in}}$ step historical data $\mathbf{X}$:
% \vspace{-5pt}
\begin{equation}
        \left[\mathbf{X}_{t-T_{\text{in}}+1: t},\mathcal{G}\right] \stackrel{f}{\rightarrow} \hat{\mathbf{X}}_{t+1:t+T_{\text{out}}}   
\end{equation}
where $F$ is the number of traffic variables to predict.

\subsection{Time-Varying Dynamic Bayesian Network}
% outline
% - what is DBN about
% - formulation of stationary DBN
% - from DBN to TVDBN
% others:
% - add a figure about DBN and TVDBN
We start with DBN that captures stationary spatiotemporal causal dependencies between time series. The value of the $i$-th time series at time $t$ is represented by $\mathbf{x}_{i,t}\in \mathbb{R}^D$. We assume that time series influence each other in both a contemporaneous and a time-lagged manner, which are called intra-slice and inter-slice dependencies, respectively. Such dependencies can be determined by a graph structure $\mathcal{B}$ and a set of parameterized functions $f_{i,t}$. If the value of $x_{j,t-k}$ affects the value of $x_{i, t}$, the time series $j$ belongs to the $k$-lag parent set of the time series $i$, denoted by $\pi_i^k$. We model the dependencies using the following general structural equation model (SEM):
% \vspace{-1pt}
\begin{equation}
        \label{eq:def_sem}
        \mathbf{x}_{i,t}=f_i(\mathbf{X}_{\pi_i^0, t},...,\mathbf{X}_{\pi_i^K, t-K}; \mathcal{B}),
\end{equation}
where $K$ is the lag order and $\mathbf{X}_{\pi_i^k, t-k}\in \mathbb{R}^{|\pi_i^k|\times D}$ represents the feature tensor of the $k$-lag parent of the time series $i$ at time $t$. For the linear condition, the SEM follows the standard structural VAR model \cite{kilian2013structural}:
% \vspace{-1pt}
\begin{equation}
    \label{eq:svar_vec}\mathbf{x}_{i,t}=\sum_{k=0}^K\sum_{i\in\pi_i^k}b_{ij}^k\cdot\mathbf{x}_{j,t-k}+\mathbf{z}_{i,t},
    % \vspace{-5pt}
\end{equation}
where $b_{ij}^k$ is the influence coefficient that captures the directional relationship of the time series $j$ to the time series $i$ with lag $k$. $\mathbf{z}_{i,t}$ is a vector of centered error terms that are independent within and across time. $b_{ij}^k$ is the element of the $k$-lag weighted adjacency matrices of $\mathcal{B}$, which allows us to write the equation \eqref{eq:svar_vec} in matrix form:
% \vspace{-1pt}
\begin{equation}
        \label{eq:svar_mat}
        \mathbf{X}_t=\sum_{k=0}^K\mathbf{B}^k\mathbf{X}_{t-k}+\mathbf{Z}_t,
        % \vspace{-1pt}
\end{equation}
where $\mathbf{B}^k=(b_{ij}^k)$ and $b_{ij}^k=0$ if $i\not\in\pi_{j}^k$. 
The directed graph represented by intra-slice matrix $\mathbf{B}^0$ is required to be acyclic in causal meaning.

\begin{figure}[t]
        \centering
        \includegraphics[width=0.85\linewidth]{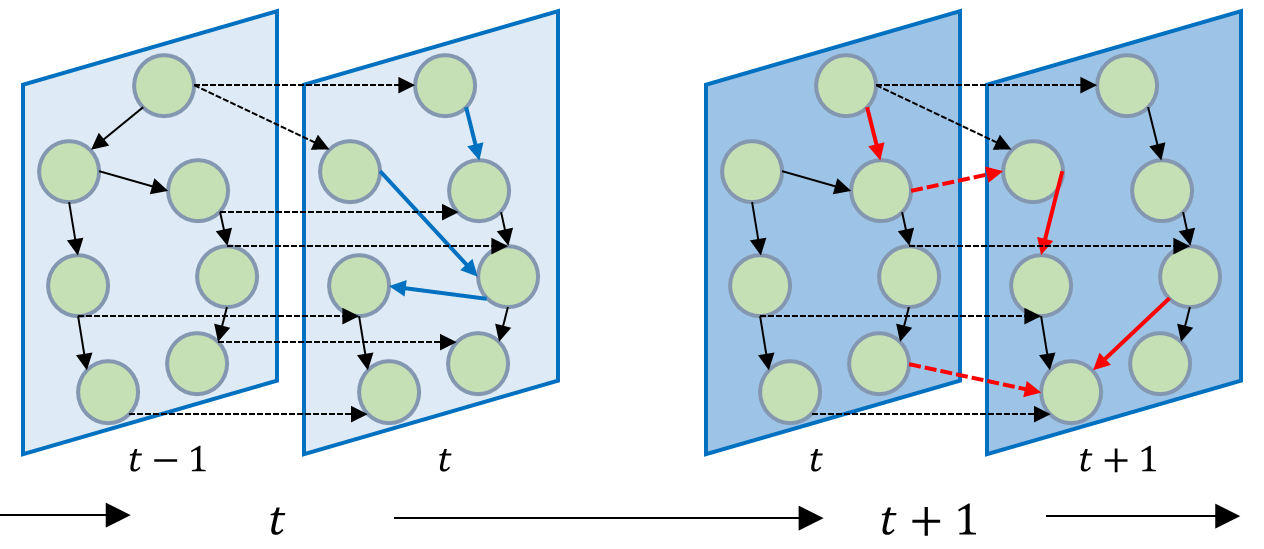}\
        \caption{Example of time-varying DBN with 7 nodes and lag $K=1$.}
        \label{fig:tvdbn}
        % \vspace{-15pt}
\end{figure}

Traditional DBN assumes a stationary data generation process, i.e., $\{\mathbf{B}^k\}_{k=0}^K$ remains the same for each $t$, which is not always valid in reality. As in Fig. \ref{fig:tvdbn}, time-varying DBN allows the graph structure and parameters to change over time. The SEM of TVDBN is:
% \vspace{-1pt}
\begin{equation}
        \mathbf{x}_{i,t}=f_{i,t}(\mathbf{X}_{\pi_{i,t}^0, t},...,\mathbf{X}_{\pi_{i,t}^K, t-K};\mathcal{B}_t)
\end{equation}
with $\mathcal{B}_t=\{\mathcal{B}^k_t\}_{k=0}^K$ and the corresponding linear SEM is:
% \vspace{-2pt}
\begin{equation}
        \label{eq:li_sem}
        \mathbf{X}_t=\sum_{k=0}^K\mathbf{B}^k_t\mathbf{X}_{t-k}+\mathbf{Z}_t
\end{equation}

\section{METHODOLOGY}
% 2 page
\begin{figure*}[thbp]
        \centering
        \includegraphics[width=1\linewidth]{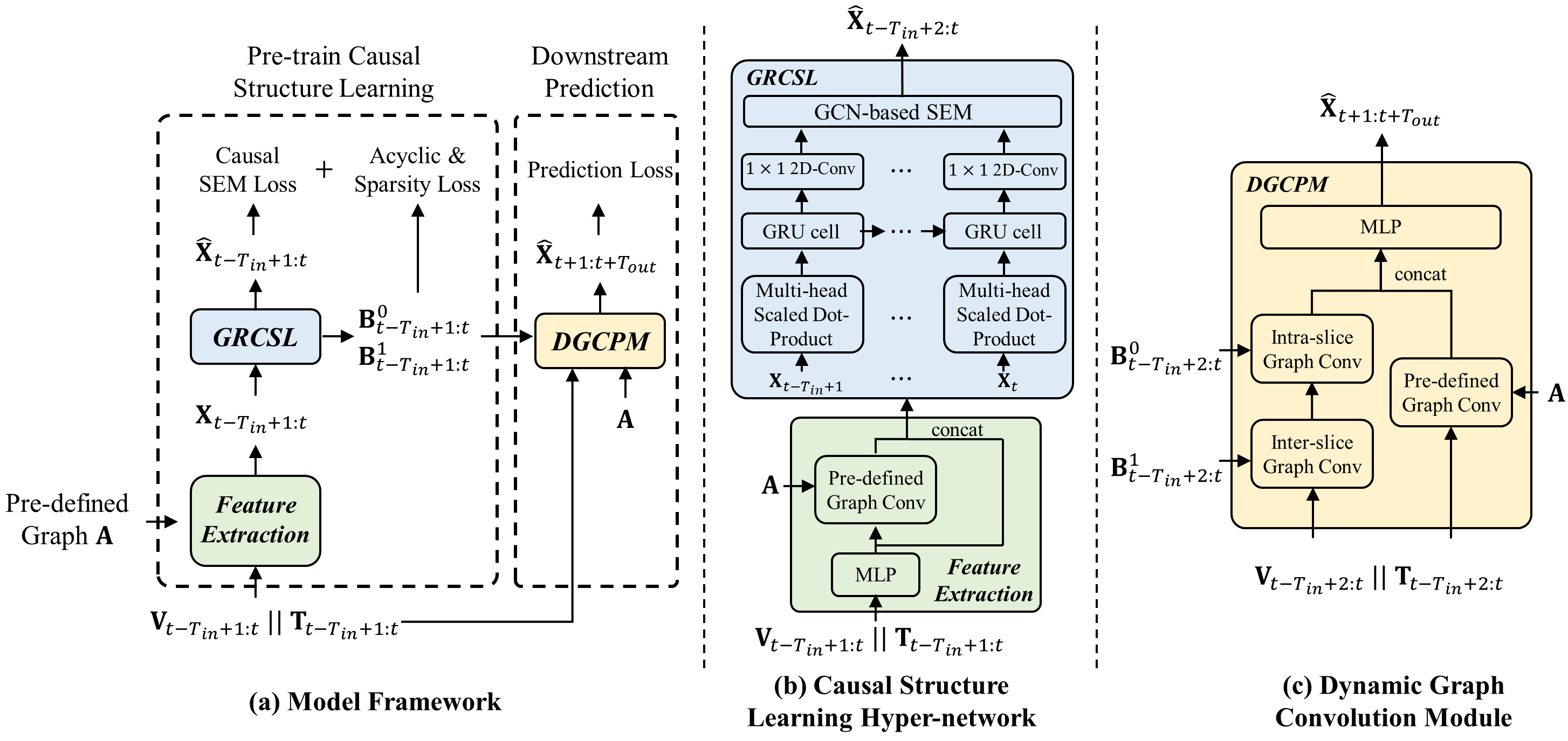}
        \vspace{-10pt}
        \caption{Architecture of the DCGCN. (a) shows the overall framework of the DCGCN. (b) shows the details of the pre-trained causal learning hyper-network, including the feature extraction block to incorporate prior information and the GCN-based recurrent causal structure learning (GRCSL) module. (c) shows the details of the dynamic graph convolution prediction module (DGCPM) used to generate traffic forecasts. %\lin{Slight changes on the figure to better distinguish between the main body of the architecture and each module}
        }
        \label{fig:framework}
        % \vspace{-15pt}
\end{figure*}

\subsection{Model Framework}
% outline
% - main components of TVDBGCN
% - ituition of each components 
In this section, we introduce the architecture of the proposed \textbf{D}ynamic \textbf{C}ausal \textbf{G}raph \textbf{C}onvolutional \textbf{N}etwork (DCGCN). As shown in Fig. \ref{fig:framework}, DCGCN is composed of three blocks including feature extraction, GCN-based recurrent causal structure learning (GRCSL) and dynamic graph convolution prediction module (DGCPM). The feature extraction block incorporates time of day and prior graph structures to augment the traffic data. The resulting features are fed into GRCSL to generate graphs of TVDBN. GRCSL is trained as a hyper-network based on the causal reconstruction loss of the input sequence. DGCPM applies graph convolution to the dynamic causal graphs of TVDBN generated by the pre-trained GRCSL and produces the final traffic forecasts. The details of each block are explained in the following subsections.

\subsection{Feature Extraction}

To accurately model traffic conditions, it is crucial to consider the dynamic spatiotemporal dependencies of the traffic network, which are influenced by the topology of the network and exhibit diverse characteristics on short and long-term time horizons. Therefore, it is necessary to integrate geographic and temporal information into the input features to describe the dynamic traffic status.
% For example, the temporal correlations between the morning and afternoon of the same day may be very different but the temporal correlations at the same time on different days may be very similar.

At each discrete time step $t$, $\mathbf{V}_t$ is the value of traffic parameters (speed or flow) and $\mathbf{T}_t$ represents time of day. Given the weighted adjacency matrix $\mathbf{A}$ of geographical traffic network as prior information, we extract the spatial features by spectral-based graph convolution \cite{kipf2016semi}:
\begin{equation}
        \label{eq:fe_geo}
        \mathbf{S}_t=\text{GCONV}^{1}_L(\mathbf{V}_t,\mathbf{A}),
\end{equation}
where $\text{GCONV}^1_L(\mathbf{X},\mathbf{A})$ is a $L$ layers spectral-based graph convolution operation with skip connection defined as
\vspace{-2pt}
\begin{equation}
        \label{eq:def_gconv}
        \begin{aligned}
                &\text{GCONV}^1_L(\mathbf{X},\mathbf{A})=\mathbf{H}^{(L)},\\
                &\mathbf{H}^{(0)}=\mathbf{X}\mathbf{\Theta}^{(0)},\ \mathbf{H}^{(l)}=\text{ReLU}(\hat{\mathbf{A}}\mathbf{H}^{(l-1)}\mathbf{\Theta}^{(l)})+\mathbf{H}^{(l-1)},\\
                &\tilde{\mathbf{A}}=\mathbf{A}+\mathbf{I},\ \tilde{\mathbf{D}}_{ii}=\sum_j\tilde{\mathbf{A}}_{ij},\ \hat{\mathbf{A}}=\tilde{\mathbf{D}}^{-\frac{1}{2}}\tilde{\mathbf{A}}\tilde{\mathbf{D}}^{\frac{1}{2}},
        \end{aligned}
\end{equation}
with adjacency matrix $\mathbf{A}$, identity matrix $\mathbf{I}$ and learnable parameters $\mathbf{\Theta}^{(l)}$ for $l=1,...,L$. The final features are obtained by concatenation
\begin{equation}
        \mathbf{X}_t=\mathbf{V}_t\ \|\ \mathbf{T}_t\ \|\ \mathbf{S}_t.
\end{equation}
\revise{where concatenation operator $||$ represents joining two matrices or tensors along the last dimension.} 

\subsection{GCN-Based Recurrent Causal Structure Learning}
% GRCSL

% Pixel-wise graph generation, gumbel-sigmoid activation
% dot-product, pixel-wise, matrix GRU, generation network, why K=2

We propose to capture the time-varying causal structure in traffic series by RNN, which generates the dynamic causal graphs stepwise from its hidden state. The challenge is to construct suitable input features for RNN that contain information about the dynamics of the causal structure. To address this challenge, we use an attention mechanism with multi-head scaled dot-product \cite{vaswani2017attention}, to describe the spatiotemporal correlations between nodes, i.e.,
% \vspace{-4pt}
\begin{equation}
        \begin{aligned}
                &\text{sdot}(\mathbf{Q}, \mathbf{K})=\frac{(\mathbf{Q}\mathbf{W_Q})(\mathbf{K}\mathbf{W}_K)^\top}{\sqrt{d}}\in\mathbb{R}^{N\times N},\\
                &\text{msdot}(\mathbf{Q}, \mathbf{K})=\left[\text{sdot}_1(\mathbf{Q}, \mathbf{K}),...,\text{sdot}_h(\mathbf{Q}, \mathbf{K})\right]\in\mathbb{R}^{N\times N\times h},\\
                &\tilde{\mathbf{C}}^0_t=\text{msdot}(\mathbf{X_t},\mathbf{X}_t),\\
                &\tilde{\mathbf{C}}^k_t=\text{msdot}(\mathbf{X_t},\mathbf{X}_{t-k}),\ k=1,...,K,
        \end{aligned}
\end{equation}
where $\mathbf{W}_Q$ and $\mathbf{W}_K$ are learnable parameters of linear transformation, $h$ is the number of heads and the learnable parameters of each head are different. $\tilde{\mathbf{C}}^0_t$ and $\tilde{\mathbf{C}}^k_t$ are indicators of intra-slice and $k$-lag inter-slice spatial correlations, respectively. We further flatten the tensors $\{\tilde{\mathbf{C}}_t^k\}_{k=0}^K$ to matrices in $\mathbb{R}^{N^2\times h}$, denoted by $\{\mathbf{C}_t^k\}_{k=0}^K$, such that each row of $\mathbf{C}_t^k$ captures the dynamics of an element in $\mathbf{B}_t^k$. Subsequently, $\mathbf{C}_t^k$ is treated as the input of GRU, i.e.,
\begin{equation}
        \begin{aligned}
                \mathbf{R}_t^k=&\sigma(\mathbf{C}_t^k\mathbf{W}_{cr}^{k}+\mathbf{H}_{t-1}^k\mathbf{W}_{hr}^{k}+\mathbf{b}_r^k),\\
                \mathbf{Z}_t^k=&\sigma(\mathbf{C}_t^k\mathbf{W}_{cz}^{k}+\mathbf{H}_{t-1}^k\mathbf{W}_{hz}^{k}+\mathbf{b}_z^k),\\
                \tilde{\mathbf{H}}_t^k=&\tanh(\mathbf{C}_t^k\mathbf{W}_{ch}^{k}+(\mathbf{R}_t^k\odot\mathbf{H}_{t-1}^k)\mathbf{W}_{hh}^{k}+\mathbf{b}_h^k),\\
                \mathbf{H}_t^k=&\mathbf{Z}_t^k\odot\mathbf{H}_{t-1}+(1-\mathbf{Z}_t^k)\odot\tilde{\mathbf{H}}_t^k,
        \end{aligned}
\end{equation}
where $\mathbf{W}_{cr}^{k},\mathbf{W}_{cz}^{k},\mathbf{W}_{ch}^{k},\mathbf{W}_{hr}^{k},\mathbf{W}_{hz}^{k},\mathbf{W}_{hh}^{k},\mathbf{b}_r^k$ and $\mathbf{b}_z^k,\mathbf{b}_h^k$ are learnable parameters. \revise{$\odot$ represents element-wise product between two matrices or tensors.} $\mathbf{H}_t^k\in\mathbb{R}^{N^2\times H}$ is the hidden state with dimension $H$. Then we unflatten the hidden states as tensors in $\mathbb{R}^{H\times N\times N}$ and generate causal graphs by three layers $1\times1$ convolution:
\begin{equation}
        \tilde{\mathbf{B}}_t^k=\text{CONV}_{1\times1}(\text{Unflat}(\mathbf{H}_t^k))\in\mathbb{R}^{N\times N},
\end{equation}
where ReLU is added after each layer except the last. Following MCSL, we apply Gumbel-Sigmoid to force the elements of $\tilde{\mathbf{B}}_t^k$ to be close to $0$ or $1$.
% and threshold the matrices at 0.5 to eliminate insignificant elements. 
In addition, the diagonal elements of the intra-slice matrices are masked as zero to satisfy the acyclic constraint of $\mathbf{B}_t^0$.
% \begin{equation}
%         \begin{aligned}
%                 &\mathbf{B}_t^k=\text{Gumbel-Sigmoid}(\tilde{\mathbf{B}}_t^k),\ k=0,...,K\\
%                 &\text{diag}(\mathbf{B}_t^0)=0
%         \end{aligned}
% \end{equation}

Considering the complex nonlinear dynamics in real traffic data, we extend the linear SEM Eq. \eqref{eq:li_sem} to a nonlinear version by spatial-based graph convolution \cite{hamilton2017inductive}, i.e.,
% \vspace{-2pt}
\begin{equation}
        \label{eq:gcn_sem}
        \mathbf{X}_t=\sum_{k=0}^K\text{GCONV}^2_L(\mathbf{X}_{t-k}, \mathbf{B}^k_t)+\mathbf{Z}_t,
\end{equation}
where $\text{GCONV}^2_L$ is a $L$ layers spatial-based graph convolution operation with skip connections defined as:
% \vspace{-2pt}
\begin{equation}
        \label{eq:def_spa_gconv}
        \begin{aligned}
                &\text{GCONV}^2_L(\mathbf{X},\mathbf{A})=\mathbf{H}^{(L)},\\
                &\mathbf{H}^{(0)}=\mathbf{X}\mathbf{\Theta}^{(0)},\ \mathbf{H}^{(l)}=\text{ReLU}(\hat{\mathbf{A}}\mathbf{H}^{(l-1)}\mathbf{\Theta}^{(l)})+\mathbf{H}^{(l-1)},\\
                &\mathbf{D}_{ii}=\sum_j\mathbf{A}_{ij},\ \hat{\mathbf{A}}=\mathbf{D}^{-1}\mathbf{A}.
        \end{aligned}
\end{equation}
With the GCN-based SEM, we reconstruct the current state value with the generated graphs by
% \vspace{-2pt}
\begin{equation}
        \hat{\mathbf{X}}_t=\text{MLP}\left(\sum_{k=0}^K\text{GCONV}^2_L(\mathbf{X}_{t-k}, \mathbf{B}_t^k)\right),
\end{equation}
where MLP is the multilayer perceptron.
% is it appropriate for directed graph convolution

By minimizing the sum of squares between $\hat{\mathbf{X}}_t$ and $\mathbf{X}_t$ and increasing the sparsity of $\{\mathbf{B}_t^k\}_{k=0}^K$ under the acyclicity constraint of intra-slice graphs, we learn the GCN-based recurrent casual structure learning (GRCSL) network. Note that when the input time series length is set to $T_{\text{in}}$, GRCSL can only generate $K$-lag DBNs of the last $T_{\text{in}}-K$ time steps. To balance the number of time steps for generating causal graphs and the number of lag in causal learning, both of which impact the representation capability in the DGCPM, we choose $K=1$ in this paper.

% GCN-based SEM

\subsection{Dynamic Graph Convolution Prediction Module}
Built upon GRCSL, we propose a dynamic graph convolution prediction module (DGCPM) to predict multi-step traffic status. In DGCPM, we redefine $\mathbf{X}_t=\mathbf{V}_t\ \|\ \mathbf{T}_t$ and incorporate time-varying causal structure $\mathcal{B}_{t}$ together with the static graph $\mathbf{A}$. 
%of prior static graph after dynamic graph convolution.

Note that SEM is designed to reconstruct the current state of a variable from the current and historical states of its parent variables. However, since the future state of the variable is unknown, intra-slice dependencies are not available in prediction. To address this problem, we decompose the general SEM in prediction into a two-step process:
% \vspace{-2pt}
\begin{equation}
        \begin{aligned}
                &\tilde{\mathbf{x}}_{i,t}=f_{i,t}^{\text{inter}}(\mathbf{X}_{\pi_{i,t-1}^1,t-1},...,\mathbf{X}_{\pi_{i,t-1}^K,t-K};\mathcal{B}_{t-1}^1,...,\mathcal{B}_{t-1}^K),\\
                &\hat{\mathbf{x}}_{i,t}=f_{i,t}^{\text{intra}}(\tilde{\mathbf{X}}_{\pi_{i,t-1}^0, t};\mathcal{B}_{t-1}^0),
        \end{aligned}
        % \vspace{-2pt}
\end{equation}
% \chen{or $\tilde{\mathbf{x}}_{i,t}=f_{i,t}^{\text{inter}}(\mathbf{X}_{\pi_i^1,t-1},...,\mathbf{X}_{\pi_i^K,t-K};\mathbf{B}_t^1,...,\mathbf{B}_t^K)$}
% \lin{for general SEM, we should use $\mathcal{B}$ to represent the structure of DBN instead of the specific adjacency matrix $\mathbf{B}$}
where $\tilde{\mathbf{x}}_{i,t}$ and $\hat{\mathbf{x}}_{i,t}$ are the initial and final predictions of $\mathbf{x}_{i,t}$, respectively. $f_{i,t}^{\text{inter}}$ takes historical data as input to capture inter-slice dependencies and $f_{i,t}^{\text{intra}}$ takes the initial prediction of $\tilde{\mathbf{X}}_{\pi_i^0,t}$ as input to capture intra-slice dependencies. Given $K=1$, for GCN-based SEM at the future time $t$, we have:
% \vspace{-2pt}
\begin{equation}
        \label{eq:two_step_gcon}
        \begin{aligned}
                \tilde{\mathbf{X}}_t&=\text{GCONV}^2_L(\mathbf{X}_{t-1}, \mathbf{B}^1_{t-1}),\\
                \mathbf{H}_{t-1}&=\text{GCONV}^2_L(\tilde{\mathbf{X}}_t, \mathbf{B}^0_{t-1}),
        \end{aligned}
        % \vspace{-3pt}
\end{equation}
where $(\mathbf{B}^0_{t-1}, \mathbf{B}^1_{t-1})$ is treated as an approximation to $(\mathbf{B}^0_{t}, \mathbf{B}^1_{t})$. To highlight the importance of the current state of a node in predicting its future state, we add an identity matrix to the adjacency matrices before graph convolution. We define the above two-step graph convolution process on TVDBN as dynamic graph convolution $\text{DyGCONV}_L(\mathbf{X}_{t-1},\mathbf{B}^0_{t-1}, \mathbf{B}^1_{t-1})$.

Given the dynamic causal graphs $\{(\mathbf{B}^0_t, \mathbf{B}^1_t)\}_{t=1}^{P}$ generated from GRCSL, we fuse the output features of dynamic graph convolution at all time steps with node embeddings generated from spectral-based prior graph convolution, i.e.,
% \vspace{-2pt}
\begin{equation}
        \begin{aligned}
                \mathbf{S}_t&=\text{GCONV}_L^1(\mathbf{V}_t\ \| \mathbf{A}),\ \tilde{\mathbf{H}}_t=\mathbf{H}_t\ \| \mathbf{S}_t,\\
                \mathbf{H}&=[\tilde{\mathbf{H}}_{t-T_\text{in}+2},...,\tilde{\mathbf{H}}_t]\in\mathbb{R}^{(T_\text{in}-1)\times N\times H}.
        \end{aligned}
        % \vspace{-2pt}
\end{equation}
Then we obtain the traffic prediction by applying a linear transformation to the above features:
% \vspace{-2pt}
\begin{equation}
        \label{eq:DGCPM}
        \begin{aligned}
                &\tilde{\mathbf{H}}=\text{Reshape}(\mathbf{H})\in\mathbb{R}^{(T_\text{in}-1)\times (N\times H)},\\
                &\hat{\mathbf{X}}_{t+1:t+T_\text{out}}=\text{Reshape}(\tilde{\mathbf{H}}\mathbf{W}_\text{out})\in\mathbb{R}^{T_\text{out}\times N\times F},
        \end{aligned}
        % \vspace{-2pt}
\end{equation}
where $\mathbf{W}_\text{out}$ is the learnable parameter of linear transformation.

\subsection{Model Training}
% outline: 
% 1. independent training for graph gen and prediction
% 2. loss function
% 3. dual lagrangian to ensure acyclic constraints
% 4. curriculum learning 
To reduce the computation complexity, we train GRCSL and DGCPM sequentially. In particular, GRCSL combined with the feature extraction block, is treated as a hyper-network and trained by optimizing the following problem:
\begin{equation}
        \label{eq:grcsl}
        \begin{aligned}
                \min_{\Theta_{c}}\quad f(\Theta_c)=&\frac{1}{T_\text{in}-1}\sum_{t=2}^{T_\text{in}}\big[\frac{1}{2}\|\hat{\mathbf{X}}_t-\mathbf{X}_t\|_2^2+\\
                &\lambda\left(\|\mathbf{B}^0_t\|_1+\|\mathbf{B}^1_t\|_1\right)\big],\\  
                \text{s.t.}\quad h(\mathbf{B}^0_t)=&\text{tr}(e^{\mathbf{B}^0_t\odot\mathbf{B}^0_t})-N=0,\ t=2,...,T_\text{in},
        \end{aligned}
\end{equation}
where $\Theta_c$ represents all the learnable parameters of GRCSL, $\|\mathbf{A}\|_1=\sum_i\sum_j|A_{ij}|$, $\|\mathbf{A}\|_2=\sqrt{\sum_i\sum_j\mathbf{A}_{ij}^2}$, $h(\mathbf{A})=0$ is the NOTEARS constraint to ensure the acyclicity of intra-slice graphs, and $\text{tr}(\cdot)$ means the trace of a matrix. Note that we ignore the offset in time index here to improve readability. We use the augmented Lagrangian approach to solve problem Eq. \eqref{eq:grcsl} with the following augmented Lagrangian function\cite{yu2019dag}:
% \vspace{-5pt}
\begin{equation}
        L_\rho(\Theta_c,\alpha)=f(\Theta_c)+\alpha \sum_{t=2}^{T_\text{in}}|h(\mathbf{B}_t^0)|+\frac{\rho}{2}\left(\sum_{t=2}^{T_\text{in}}|h(\mathbf{B}^0_t)|\right)^2,
\end{equation}
where $\rho>0$ is the step size parameter and $\alpha$ is the Lagrange multiplier. When $\rho$ goes to positive infinity, the minimizer of $L_\rho(\Theta_c,\alpha)$ must satisfy the NOTEARS constraint. We update $\Theta_c$, $\alpha$ and $\rho$ with the following strategy:
\begin{align}
        \label{eq:lagr_sub}
        \Theta_c^{k+1}=&\underset{\Theta_c}{\operatorname{argmin}}L_{\rho^k}(\Theta_c,\mu^k),\\
        \alpha^{k+1}=&\alpha^k+\rho^k\sum_{t=2}^{T_\text{in}}|h(\mathbf{B}_t^0)|,\\
        \rho^{k+1}=&\begin{cases}
                \eta\rho^k,&\text{if $\left[\sum_{t=2}^{T_\text{in}}|h(\mathbf{B}_t^0)|\right]_{k+1}>\gamma \left[\sum_{t=2}^{T_\text{in}}|h(\mathbf{B}_t^0)|\right]_{k}$},\\
                \rho^k,&\text{o.w.}
        \end{cases}
\end{align}
where $k$ is the iteration index, $\eta > 1$ and $0<\gamma< 1$ are hyperparameters. We stop iterations when $\sum_{t=2}^{T_\text{in}}|h(\mathbf{B}_t^0)|$ is less than a small number $\xi$. The subproblem Eq. \eqref{eq:lagr_sub} is to learn GRCSL with $L_{\rho^k}(\Theta_c,\mu^k)$ as the loss function, which can be solved by stochastic gradient descent.

Built upon the pre-trained GRCSL, we train the DGCPM by using the mean absolute error loss as the objective function and minimizing the multi-step prediction loss with curriculum learning. We start by training the model to make one-step predictions and gradually increase the prediction length to its maximum $T_\text{out}$ as the training iterations progress. This approach helps the algorithm to find a good local minimum for the model in the early stages, which is advantageous for achieving better prediction performance in the end.

\section{EXPERIMENTS}

\begin{table}[t]
        \centering
        \caption{Traffic Forecasting Performance Comparison for METR-LA}
        \label{tab:baselines}
        \begin{threeparttable} 
                \setlength{\tabcolsep}{1.2mm}{
                \scalebox{0.82}{
                \begin{tabular}{cccccccccc}
                \toprule
                & \multicolumn{3}{c}{Horizon 3} & \multicolumn{3}{c}{Horizon 6} & \multicolumn{3}{c}{Horizon 12} \\
                Methods & MAE & MAPE & RMSE & MAE & MAPE & RMSE & MAE & MAPE & RMSE \\
                \midrule
                VAR & 4.42 & 10.20\% & 7.89 & 5.41 & 12.70\% & 9.13 & 6.52 & 15.80\% & 10.11 \\
                DCRNN & 2.77 & 7.30\% & 5.38 & 3.15 & 8.80\% & 6.45 & 3.60 & 10.50\% & 7.6 \\
                STGCN & 2.88 & 7.62\% & 5.74 & 3.47 & 9.57\% & 7.24 & 4.59 & 12.70\% & 9.4 \\
                ASTGCN & 3.09 & 8.06\% & 5.47 & 3.67 & 9.87\% & 6.46 & 4.43 & 12.57\% & 7.64 \\
                AGCRN & 2.87 & 7.70\% & 5.58 & 3.23 & 9.00\% & 6.58 & 3.62 & \textbf{10.38\%} & 7.51 \\
                % MTGNN & 2.69 & 6.86\% & 5.18 & 3.05 & 8.19\% & 6.17 & 3.49 & 9.87\% & 7.23 \\
                % DGCRN & 2.72 & 6.99\% & 5.21 & 3.10 & 8.42\% & 6.24 & 3.55 & 10.23\% & 7.35  \\
                \textbf{DCGCN} & \textbf{2.73} & \textbf{7.04\%} & \textbf{5.23} & \textbf{3.11} & \textbf{8.62\%} & \textbf{6.28} & \textbf{3.57} & 10.47\% & \textbf{7.41}  \\
                \midrule
                $\text{DBGCN}_{20}$ & 2.79  & 7.29\% & 5.13  & 3.16  & 8.60\%  & 6.04  & 3.69  &  10.34\% & 7.14 \\
                $\text{\textbf{DCGCN}}_{20}$ &  \textbf{2.72}  & \textbf{7.06\%} & \textbf{5.01}  & \textbf{3.05}  & \textbf{8.37\%}  & \textbf{5.92}  & \textbf{3.48}  & \textbf{9.94\%}  & \textbf{6.94}\\
                \bottomrule
                \end{tabular}
                }}
                \begin{tablenotes}
                        \scriptsize
                        \item[1] Methods with subscript 20 runs on a sub-dataset with 20 nodes.
                \end{tablenotes}
        \end{threeparttable}
        % \vspace{-15pt}
\end{table}

% 1 page
\subsection{Experimental Settings}
We conduct experiments on a public real-world traffic dataset \textbf{METR-LA}, which contains traffic speed data collected from loop detectors in the highway of Los Angeles \cite{jagadish2014big}. METR-LA collects 4 months of data ranging from Mar 1st 2012 to Jun 30th 2012 from 207 selected sensors. In our experiments, the sensors in different locations are viewed as nodes in the traffic network.

% pre-process: 1. normalization, zero mask
Following DCRNN \cite{li2017diffusion}, we set the time granularity of speed data to 5 minutes and apply Z-Score normalization. 70\% of data is used for training, 20\% are used for testing while the remaining 10\% for validation. The predefined weighted adjacency matrix $\mathbf{A}$ is constructed by calculating the pairwise road network distances between sensors:
% \vspace{-3pt}
\begin{equation}
        \mathbf{A}_{ij}=
        \begin{cases}
                \exp\left(-\frac{\operatorname{dist}\left(v_i, v_j\right)^2}{\sigma^2}\right),&\text{if $\text{dist}(v_i,v_j)\leq \kappa$}\\
                0,&\text{o.w.}
        \end{cases}
        % \vspace{-2pt}
\end{equation}
where $\text{dist}(v_i,v_j)$ is the road network distance from sensor $v_i$ to sensor $v_j$. $\sigma$ is the standard deviation of distances and $\kappa=0.1$ is the threshold parameter. The input and output sequence length $T_\text{in}$ and $T_\text{out}$ are set to $12$. The prediction target is traffic speed with dimension $F=1$. Number of layers in graph convolution $L$ is $4$. For missing values marked as zero in METR-LA, we filter out the missing values in loss function and metrics calculation.

\revise{Following previous work\cite{yu2019dag}, we empirically set the hyperparameters $\tau=0.2$, $\eta=10$ and $\gamma=0.5$. The $l_1$ penalty weight $\lambda$ is chosen as $2\times 10^{-5}$ to slightly control false discoveries. A detailed selection of the hyperparameters could also be achieved by a grid search, but this involves high computational costs. The initial value of the Lagrange multiplier is set to $0$, i.e. $\alpha^0=0$. And $\rho^0$ is set to $10^{-3}$ so that the optimization program focuses on minimizing the reconstruction error at an early stage.}

\subsection{Baselines and Evaluation Metrics}
% too wordy
% (1) \textbf{VAR}: a classical time series analysis model which captures linear correlations between future traffic flow and historical data; (2)\textbf{ARIMA} a classical statictical model which integrates auto-regression with moving average model; 
We compare our proposed method with the classical and state-of-the-art deep learning based models, including (1) \textbf{VAR} \cite{zivot2006vector}: a statistical model which captures linear correlations between future traffic series and historical data; (2) \textbf{DCRNN} \cite{li2017diffusion}: an RNN-based model which integrates diffusian graph convolutions into RNN with encoder-decoder structure; (3) \textbf{STGCN} \cite{yu2017spatio}: a complete convolutional model which capture spatial and temporal dependencies by graph convolution and 1D-convolution on time axis, respectively; (4) \textbf{ASTGCN} \cite{guo2019attention}: adding an attention mechanisms into STGCN by using spatial and temporal attention to adjust the input before convolution. (5) \textbf{AGCRN} \cite{bai2020adaptive}: an RNN-based model which learns an adaptive static graph and integrates graph convolution with node adaptive parameter learning into GRU gate; 
% (5) \textbf{MTGNN} \cite{wu2020connecting}: a convolutional model which learns an adaptive static directed graph and employs mix-hop propagation and dilated inception layer in spatial and temporal convolution modules, respectively; 
% (6) \textbf{DGCRN} \cite{li2021dynamic}: an RNN-based model which generates dynamic adjacency matrix from hidden states of RNN and integrates mix-hop dynamic graph convolution into GRU gate. 
(6) \textbf{DBGCN} \cite{luan2022traffic}: an GCN-based model that learns an adptive stationary DBN through statistical methods and applies graph convolution sequentially to graphs of the DBN. This is the only work also considering causality for spatial dependence analysis.

For all the baselines, we use the default settings from their original papers. The performance of traffic prediction is measured by three commonly used metrics, including (1) Mean Absolute Error (\textbf{MAE}), (2) Root Squared Error (\textbf{RMSE}), and (3) Mean Absolute Percentage Error (\textbf{MAPE}). 

\begin{figure*}[t]
        % \vspace{-10pt}
        \centering
        \includegraphics[width=0.8\textwidth]{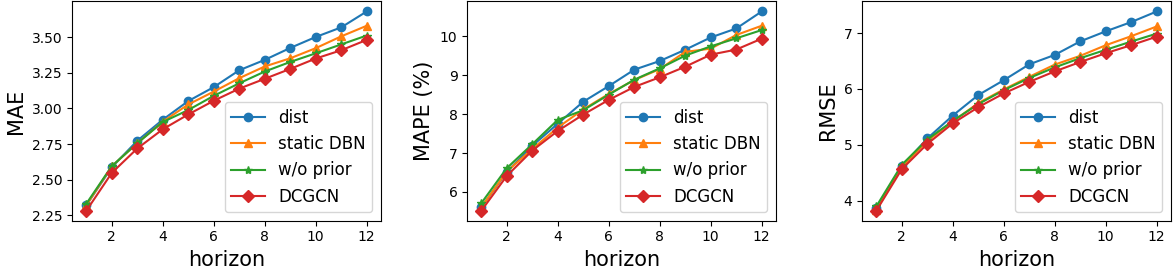}
        % \vspace{-15pt}
        \caption{Prediction performance of ablation study at each horizon.}
        \label{fig:ablation}
        % \vspace{-15pt}
\end{figure*}

\subsection{Performance Comparison}

We compare the prediction performances for the future step 3 (15 minutes), step 6 (30 minutes), and step 12 (1 hour). The comparison between all methods is shown in Table \ref{tab:baselines}. Note that since the computational cost of learning DBN for DBGCN on a large dataset is intolerable, we compare the proposed method with DBGCN on a subset of METR-LA with 20 nodes. The results show that: (1) methods with adaptive spatial dependence graphs, such as AGCRN, generally achieve better performance than other methods based on a static predefined graph, indicating the importance of adaptive spatial topologies. (2) Our methods outperform other adaptive dynamic graphs based methods in both short-term and long-term traffic prediction. This illustrates the advantages of using causality-based spatial dependence analysis in prediction compared with correlation-based analysis. (3) The two points above can also be validated by comparing our DCGCN with DBGCN, which uses static causal graph for spatial dependence. It demonstrates the efficiency and effectiveness of our proposed DCGCN. 

\subsection{Ablation Study}

% \begin{table}[t]
%         \centering
%         \caption{Ablation Study}
%         \label{tab:abalation}
%         \setlength{\tabcolsep}{1.2mm}{
%         \scalebox{0.8}{
%         \begin{tabular}{cccccccccc}
%         \toprule
%         & \multicolumn{3}{c}{Horizon 3} & \multicolumn{3}{c}{Horizon 6} & \multicolumn{3}{c}{Horizon 12} \\
%         Methods & MAE & MAPE & RMSE & MAE & MAPE & RMSE & MAE & MAPE & RMSE \\
%         \midrule
%         dist & 2.77 & 7.19\% & 5.11 & 3.15 & 8.72\% & 6.16 & 3.68 & 10.64\% & 7.39 \\
%         static DBN & 2.76 & 7.10\% & 5.07 & 3.12 & 8.52\% & 5.99 & 3.58 & 10.27\% & 7.12 \\
%         w/o prior & 2.76 & 7.23\% & 5.09 & 3.09 & 8.50\% & 5.98 & 3.51 & 10.16\% & 6.99 \\
%         DCGCN & 2.72  & 7.06\% & 5.01  & 3.05  & 8.37\%  & 5.92  & 3.48  & 9.94\%  & 6.94  \\
%         \bottomrule
%         \end{tabular}
%         }}
%         % \vspace{-15pt}
% \end{table}

To further demonstrate the effectiveness of the key modules of our DCGCN, we conduct an ablation study at METR-LA. Considering the computational burden of certain models, we perform the ablation experiments on a subset of METR-LA with 20 nodes. We name DCGCN without certain modules as follows: (1) \textbf{dist}: DCGCN without GRCSL. We replace the dynamic causal graphs generated by GRCSL with a static predefined distance graph. (2) \textbf{static DBN}: DCGCN without GRCSL. We replace the dynamic causal graphs from GRCSL with a static DBN learned using classical casuality discovery methods \cite{trabelsi2013dynamic}.
(3) \textbf{w/o prior}: DCGCN without incorporating prior information (i.e., the predefined distance graph) in feature extraction module and DGCPM.

We report the results in Fig. \ref{fig:ablation}, and can draw some conclusions as follows: (1) Compared with the predefined distance graph, DGCPM based on causal graphs including static DBN and TVDBN shows better prediction performance. It validates that causal discovery is helpful for constructing finer spatial-temporal topology representation of the traffic network. (2) DCGCN outperforms models based on static DBN, demonstrating the need to capture nonlinear and time-varying causal traffic propagation patterns. (3) Incorporating prior information such as distance-based spatial topology of the traffic network helps to construct better dynamic causal graphs, leading to better prediction performance.

\section{CONCLUSIONS}
% 1/4 page
In this work, we propose a novel approach to enhance the performance of GCN-based methods for traffic prediction by incorporating time-varying dynamic causal graphs that can capture time-varying spatiotemporal relationships of traffic series. By leveraging deep learning to model the generation mechanism of causal graphs, we can significantly reduce the computational cost of structure learning for TVDBN. Our experimental results on the METR-LA dataset show that our proposed method achieves superior prediction performance in both the short-term and long-term scales. Overall, our approach offers a promising solution for accurately modeling time-varying spatiotemporal dependencies in traffic prediction with the help of causal discovery. Further investigation into the dynamic causal graphs learned from GRCSL can be beneficial in uncovering the time-varying traffic propagation patterns in the future. Such insights can help us better understand the underlying spatial dependence relationships of traffic systems.

\addtolength{\textheight}{-5cm}   % This command serves to balance the column lengths

%%%%%%%%%%%%%%%%%%%%%%%%%%%%%%%%%%%%%%%%%%%%%%%%%%%%%%%%%%%%%%%%%%%%%%%%%%%%%%%%

%%%%%%%%%%%%%%%%%%%%%%%%%%%%%%%%%%%%%%%%%%%%%%%%%%%%%%%%%%%%%%%%%%%%%%%%%%%%%%%%



\section*{ACKNOWLEDGMENT}

This paper was supported by the NSFC Grant 72271138 and 71932006, and the BNSF Grant 9222014. This work is also partly funded by SenseTime-Tsinghua Research Collaboration Funding. 

%%%%%%%%%%%%%%%%%%%%%%%%%%%%%%%%%%%%%%%%%%%%%%%%%%%%%%%%%%%%%%%%%%%%%%%%%%%%%%%%

% \begin{thebibliography}{99}

% \bibitem{c1} G. O. Young, Synthetic structure of industrial plastics (Book style with paper title and editor), 	in Plastics, 2nd ed. vol. 3, J. Peters, Ed.  New York: McGraw-Hill, 1964, pp. 1564.

% \end{thebibliography}

\bibliographystyle{IEEEtran}
\bibliography{main}

\end{document}